\documentclass[preprints,article,moreauthors,pdftex]{Definitions/mdpi} 

\usepackage[ruled,vlined]{algorithm2e}

\firstpage{1} 
\pubvolume{1}
\issuenum{1}
\articlenumber{0}
\pubyear{2021}
\copyrightyear{2020}
\datereceived{31 May 2021} 
\dateaccepted{6 July 2021} 

\Title{\textls[-25]{Fast Flow Reconstruction via Robust Invertible $n \times n$ Convolution}}

\TitleCitation{Fast Flow Reconstruction via Robust Invertible $n \times n$ Convolution}

\Author{Thanh-Dat Truong $^{1,*}$, Chi Nhan Duong $^{2}$, Minh-Triet Tran $^{3}$, Ngan Le $^{1}$ and Khoa Luu $^{1}$} 

\AuthorNames{Thanh-Dat Truong Chi Nhan Duong, Minh-Triet Tran, Ngan Le and Khoa Luu}

\AuthorCitation{Truong, T.-D.; Duong, C.N.; Tran, M.-T.; Le, N.; Luu, K.}

\address{%
$^{1}$ \quad Computer Vision and Image Understanding Lab,\\ 
\quad\; Department of Computer Science and Computer Engineering, \\
\quad\; University of Arkansas, Arkansas, USA; \\
\quad\; thile@uark.edu (N.L.); khoaluu@uark.edu (K.L.)\\%
$^{2}$ \quad Department of Computer Science and Software Engineering,\\
\quad\; Concordia University, Quebec, Canada;\\
\quad\; dcnhan@ieee.org\\
$^{3}$ \quad Faculty of Information Technology, \\
\quad\; University of Science, VNU-HCM, Ho Chi Minh, Vietnam;\\
\quad\; tmtriet@hcmus.edu.vn}

\corres{Correspondence: tt032@uark.edu; Tel.: +1-479-404-0772}

\abstract{Flow-based generative models have recently become one of the most efficient approaches to model data generation. Indeed, they are constructed with a sequence of invertible and tractable transformations. Glow 
first introduced a simple type of generative flow using an invertible $1 \times 1$  convolution. However, the $1 \times 1$ convolution suffers from limited flexibility compared to the standard convolutions. In this paper, we propose a novel invertible $n \times n$ convolution approach that overcomes the limitations of the invertible $1 \times 1$ convolution. In addition, our proposed network is not only tractable and invertible but also uses fewer parameters than standard convolutions. The experiments on CIFAR-10, ImageNet and Celeb-HQ datasets, have shown that our invertible $n \times n$ convolution helps to improve the performance of generative models significantly.}


\begin{document}


\section{Introduction} \label{sec:introduction}

{Supervised} deep learning models have recently achieved numerous breakthrough results in various applications, for example, Image Classification \cite{he2016deep, huang2017densely, sun2018fishnet}, Object Detection, \cite{liu2016ssd, redmon2016you, lin2017focal}, Face Recognition \cite{Le_JPR2015, Xu_IJCB2011, Xu_TIP2015, Luu_IJCB2011, Luu_BTAS2009, Duong_ICASSP2011, Luu_FG2011,Chen_FG2011}, Image Segmentation \cite{chen2017deeplab, long2015fully} and Generative Model \cite{Luu_CAI2011, mirza2014conditional, karras2017progressive, Duong_Long, Luu_ROBUST2008, Duong_TNVP}.  
However, these methods usually require a huge number of annotated data, which is highly expensive. In order to tackle the requirement of large annotations, generative models have become a feasible solution. 
The main objective of generative models is to learn the hidden dependencies that exist in the realistic data so that they can extract meaningful features and variable interactions to synthesize new realistic samples 
without human supervision or labeling.
Generative models can be used in numerous applications such as anomaly detection \cite{dimattia2019survey}, image inpainting \cite{yu2018generative}, data generation \cite{karras2017progressive, karras2019style}, super-resolution \cite{ledig2017photo}, face synthesis \cite{Duong_TNVP, Duong_automatic, Duong_2018}, and so forth. However, learning generative models is an extremely challenging process due to high-dimensional data.

There are two types of generative models extensively deployed in recent years, including likelihood-based methods \cite{glow, invertible_autoregressive,nice, real_nvp} and Generative Adversarial Networks (GANs)~\cite{gan}.
Likelihood-based methods have three main categories: Autoregressive models \cite{invertible_autoregressive}, variational autoencoders (VAEs) \cite{variational_autoencoder}, and flow-based models \cite{glow, real_nvp, nice}.
The flow-based generative model is constructed using a sequence of invertible and tractable transformations, the model explicitly learns the data distribution and therefore the loss function is simply a negative log-likelihood.

The flow-based model was first introduced in \cite{nice} and later extended in RealNVP \cite{real_nvp}. These methods introduced an affine coupling layer that is invertible and tractable based on Jacobian determinant.
As the design of the coupling layers, at each stage, only a subset of data is transformed while the rest is required to be fixed.
Therefore, they may be limited in flexibility. To overcome this limitation, coupling layers are alternated with less complex transformations that manipulate on all dimensions of the data. In RealNVP\cite{real_nvp}, the authors use a fixed channel permutation using fixed checkerboard and channel-wise masks. Kingma et~al. \cite{glow} simplifies the architecture by replacing the reverse permutation operation on the channel ordering with invertible $1 \times 1$ convolutions.

\begin{figure}[H]
    \includegraphics[width=0.7\columnwidth]{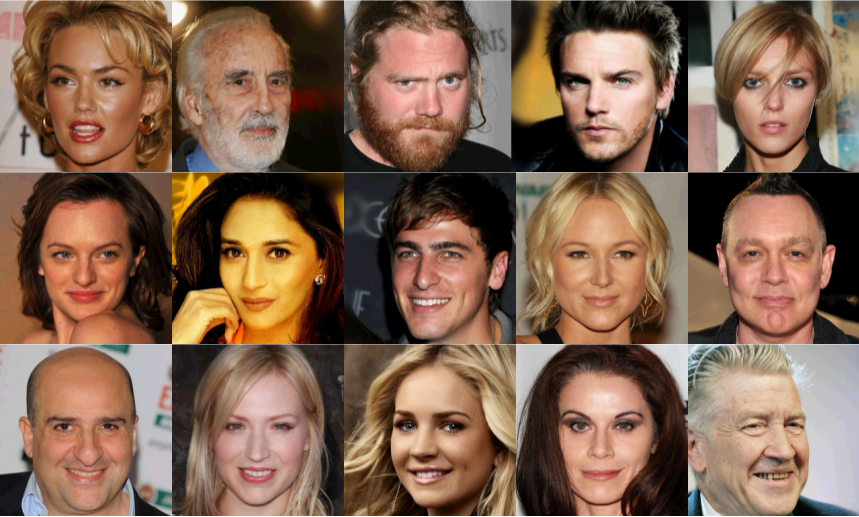}
    \caption{Reconstruction results using our proposed approach.}
    \label{fig:syn_results}
\end{figure}

However, the $1 \times 1$ convolutions are not flexible enough in these scenarios. It is extremely hard to compute the inverse form of the standard $n \times n $ convolutions, and this step usually produces high computational costs.
There are prior approaches that design the invertible $n \times n$ convolutions by using emerging convolution \cite{dxd_invertible_conv}, periodic convolutions \cite{dxd_invertible_conv}, autoregressive flow \cite{mask_autoregessive} or stochastic approximation \cite{i-resnet, inv_self_attention, residual_flows}.
In this paper, we propose an approach to generalize an invertible $1 \times 1$ convolution to a more general form of $n \times n$ convolution. Firstly, we reformulate the standard convolution layer by shifting the inputs instead of the kernels. Then, we propose an invertible shift function that is a tractable form of Jacobian determinant. Through the experiments on CIFAR-10~\cite{cifar}, ImageNet \cite{imagenet_cvpr09} and Celeb-HQ \cite{celeb_dataset} datasets, we prove that our proposals are significant and efficient for high-dimensional data.
Figure \ref{fig:syn_results} illustrates the advantages of our approach with high-resolution synthesized images.

\noindent
\textbf{Contributions:} This work generalizes the invertible $1 \times 1$ convolution to an invertible $n \times n$ convolution by reformulating the convolution layer using our proposed invertible shift function. Our contributions can be summarized as follows:

\begin{itemize}
    \item Firstly, by analyzing the standard convolution layer, we reformulate its equation into a form such that, rather than shifting the kernels during the convolution process, shifting the input provides equivalent results. 
    \item Secondly, we propose a novel invertible shift function that mathematically helps to reduce the computational cost of the standard convolution while keeping the range of the receptive fields. The determinant of the Jacobian matrix produced by this shift function can be computed efficiently.
    \item Thirdly, evaluations of several datasets on both objects and faces have shown the generalization of the proposed $n \times n$ convolution using our proposed novel invertible shift function.
    
    
\end{itemize}

\section{Related Work} \label{sec:related_work}

The generative models can be divided into two groups, that is, \textit{Generative Adversarial Networks}  and \textit{Flow-based Generative Models}. 
In the first group, \textit{Generative Adversarial Networks}~\cite{gan} provide an appropriate solution to model the data generation. The discriminative model learns to distinguish the real data from the fake samples produced using a generative model. Two models are trained as they are playing a mini-max game.
Meanwhile, in the second group, 
the \textit{Flow-based Generative Models}  \cite{glow, real_nvp, nice} are constructed using a sequence of \textbf{\textit{invertible}} and \textbf{\textit{tractable}} transformations. 
Unlike GAN, the model explicitly learns the data distribution $p(\mathbf{x})$ and therefore the loss function is efficiently employed with the log-likelihood.

In this section, we discuss several types of flow-based layers that are commonly used in flow-based generative models.
An overview of several invertible functions is provided in the Table \ref{tab:compare_function}. In particular, all functions easily obtain the reverse function and tractability of a Jacobian determinant. The symbols $\odot, /$ denote element-wise multiplication and division. $h, w$ denotes the height and width of the input/output. $c, i, j$ are the depth channel index and spatial indices, respectively.

\end{paracol}
\nointerlineskip
 \begin{specialtable}[H]

     \widetable
    \caption{Comparative invertible functions in several generative normalizing flows.}
    \begin{tabular}{llll}
\toprule
         \textbf{Description\,\,\,\,\,\,\,\,\,\,\,\,\,\,\,\,\,\,\,\,\,\,\,\,\,\,\,\,\,\,\,\,\,\,\,\,\,\,\,\,\,\,\,} &  \textbf{Function\,\,\,\,\,\,\,\,\,\,\,\,\,\,\,\,\,\,\,\,\,\,\,\,\,\,\,\,\,\,\,\,\,\,\,\,\,\,\,\,\,\,\,\,\,\,\,} & \textbf{Reverse Function\,\,\,\,\,\,\,\,\,\,\,\,\,\,\,\,\,\,\,\,\,\,\,\,\,\,\,\,\,\,} & \textbf{Log-Determinant\,\,\,\,\,\,\,\,\,\,\,\,\,} \\
         \midrule
         ActNorm \cite{glow} & $\mathbf{y} = \mathbf{x} \odot \gamma + \beta$ & $\mathbf{x} = (\mathbf{y} - \beta)/\gamma$ & $\sum \log|\gamma|$\\ 
         \midrule
         Affine Coupling \cite{real_nvp} & $\mathbf{x} = [\mathbf{x}_a,\mathbf{x}_b]$ & $\mathbf{y} = [\mathbf{y}_a,\mathbf{y}_b]$ &  $\sum \log |s(\mathbf{x}_b)|$\\
         & $\mathbf{y}_a = \mathbf{x}_a \odot s(\mathbf{x}_b) + t(\mathbf{x}_b)$ & $\mathbf{x}_a = [\mathbf{y}_a - t(\mathbf{y}_b)]/s(\mathbf{y}_b)$ &  \\ 
         & $\mathbf{y} = [\mathbf{y}_a \mathbf{x}_b]$ & $\mathbf{x} = [\mathbf{x}_a \mathbf{y}_b]$ &  \\
         \midrule
         $1 \times 1$ conv \cite{glow} & $\mathbf{y}_{:,i,j} = \mathbf{W}\mathbf{x}_{:,i,j}$ & $\mathbf{x}_{:,i,j} = \mathbf{W}^{-1}\mathbf{y}_{:,i,j}$ & $h.w.\log|\operatorname{det}\mathbf{W}|$ \\
         \midrule
         \textbf{Our Shift Function} & 
         $\mathbf{y}_{c,i,j} = \alpha_c\mathbf{x}_{c,i,j} + \beta_c$ & $\mathbf{x}_{c,i,j} = [\mathbf{y}_{c,i,j} - \beta_c]/\alpha_c$ & $h.w.\sum_c\log|\alpha_c|$ \\
    \bottomrule
    \end{tabular}
    \label{tab:compare_function}
\end{specialtable}
\begin{paracol}{2}
\switchcolumn

\textbf{Coupling Layers:} 
NICE \cite{nice} and RealNVP \cite{real_nvp} presented coupling layers with a normalizing flow by stacking a sequence of invertible bijective transformation functions. The bijective function is designed as an affine coupling layer, which is a tractable form of Jacobian determinant. RealNVP can work in a multi-scale architecture to build a more efficient model for large inputs. To further improve the propagation step, the authors applied batch normalization and weight normalization during training. Later,~Ho~et.~al.~\cite{flow_pp} presented a continuous mixture cumulative distribution function to improve the density modeling of coupling layers. In addition to improving the expressiveness of transformations of coupling layers, \cite{flow_pp} utilized multi-head self-attention layers \cite{transformer} in the transformations. 

\textbf{Inverse Autoregressive Convolution:} 
Germain et~al. \cite{made} introduced autoregressive autoencoders by constructing an extension of a non-variational autoencoder that can estimate distributions and is straightforward in computing their Jacobian determinant. Masked autoregressive flow \cite{mask_autoregessive} is a type of normalizing flow, where the transformation layer is built as an autoregressive neural network. Inverse autoregressive flow \cite{invertible_autoregressive} formulates the conditional probability of the target variable as an autoregressive model.






\textbf{Invertible $1 \times 1$ Convolution:} 
Kingma et~al.  \cite{glow} proposed simplifying the architecture via invertible $1 \times 1$ convolutions. Learning a permutation matrix is a discrete optimization that is not amenable to gradient ascent. However, the permutation operation is simply a special case of a linear transformation with a square matrix. We can pursue this work with convolutional neural networks, as permuting the channels is equivalent to a $1 \times 1$ convolution operation with an equal number of input and output channels. Therefore, the authors replace the fixed permutation with learned $1 \times 1$ convolution operations.

\textbf{Activation Normalization:} 
\cite{glow} performs an affine transformation using scale and bias parameters per channel. This layer simply shifts and scales the activations with data-dependent initialization that normalizes the activations given an initial minibatch of data. This allows the scaling down of the minibatch size to 1 (for large images) and the scaling up of the size of the model.

\textbf{Invertible $n \times n$ Convolution:} 
Since the invertible $1 \times 1$ convolution is not flexible, Hoogeboom et~al. \cite{dxd_invertible_conv} proposed an invertible $n \times n$ convolution generalized from the $1~\times~1$ convolutions. The authors presented two methods to produce the invertible convolutions: (1) \textit{Emerging Convolution} and (2) \textit{Invertible Periodic Convolutions}. %
Emerging Convolution is obtained by chaining specific invertible autoregressive convolutions \cite{invertible_autoregressive} and speeding up this layer through the use of an accelerated parallel inversion module implemented in Cython. Invertible Periodic Convolutions transform data to the frequency domain via Fourier transform; this alternative convolution has a tractable Jacobian determinant and inverse. However, these invertible $n \times n$ convolutions require more parameters; therefore, these have an additional computational cost compared to our proposed method.

\textbf{Lipschitz Constant:} 
Behrmann et~al. \cite{i-resnet} developed a theory that any residual blocks  satisfying the Lipschitz Constant can be invertible. Hence, Behrmann et~al. proposed an invertible residual network (i-ResNet) as a normalizing flow-based model. Similar to \cite{dxd_invertible_conv,real_nvp, nice, glow}, i-ResNet is learned by optimizing the negative log-likelihood in which the inverse flow and Jacobian determinant of the residual block can be efficiently approximated by the stochastic methods. Inheriting the success of Lipschitz theory, Kim et~al. \cite{inv_self_attention} proposed an $L_2$ self-attention that allows the self-attention of the Transformer networks \cite{transformer} to be~invertible.

\section{Background}

\subsection{Flow-Based Generative Model}

Let $\mathbf{x}$ be a high-dimensional vector with unknown true distribution $\mathbf{x} \sim p_{\mathcal{X}}(\mathbf{x})$, $x \in \mathcal{X}$, 
a simple prior probability distribution $p_{\mathcal{Z}}$ on a latent variable $z \in \mathcal{Z}$,
a bijection $f : \mathcal{X} \to \mathcal{Z}$, the change of variable formula defines a model distribution on $\mathcal{X}$ as shown in Equation (\ref{eqn:model_distribution}).
\begin{equation} \label{eqn:model_distribution}
    p_{\mathcal{X}}(\mathbf{x}) = p_{\mathcal{Z}}(\mathbf{z})\Big| \operatorname{det} \Big(\frac{\partial f(\mathbf{x})}{\partial \mathbf{x}}\Big) \Big|,
\end{equation}
where $\frac{\partial f(x)}{\partial x}$ is the Jacobian of $f$ at $\mathbf{x}$. The
log-likelihood objective is then equivalent to~minimizing:
\begin{equation} \label{eqn:loglikelihood}
\begin{split}
    \mathcal{L}(\mathcal{X}) &= - _{\mathbf{x} \in \mathcal{X}} \log p_{\mathcal{X}}(\mathbf{x})\\
    &= -_{\mathbf{x} \in \mathcal{X}}\Bigg[\log p_{\mathcal{Z}}(\mathbf{z}) + \log \Big| \operatorname{det} \Big(\frac{\partial f(\mathbf{x})}{\partial \mathbf{x}}\Big) \Big| \Bigg].
\end{split}
\end{equation}

Since the data $\mathbf{x}$ are discrete data, we add a random uniform noise $u \in \mathcal{U}(0, a)$, where $a$ is determined by the discretization level of the data, to make $\mathbf{x}$ be continuous data. The generative process can be defined as Equation (\ref{eqn:generative_process}).
\begin{equation} \begin{split} \label{eqn:generative_process}
    \mathbf{z} &\sim p_{\mathcal{Z}}(\mathbf{z}) \\
    \mathbf{x} &= f^{-1}(\mathbf{z}).
\end{split} \end{equation}

The bijection function $f$ is constructed from a sequence of invertible and tractable Jacobian determinant transformations: $f = f_1 \circ f_2 \circ ... \circ f_K$ ($K$ is the number of transformations). Such a sequence of invertible transformations is also called a normalizing flow. Here, Equation \eqref{eqn:loglikelihood} can be written as in Equation \eqref{eqn:loglikelihood_trans}.
\begin{equation} \label{eqn:loglikelihood_trans}
\begin{split}
    \mathcal{L}(\mathcal{X}) &= -_{\mathbf{x} \in \mathcal{X}} \log p_{\mathcal{X}}(\mathbf{x}) 
    \\&= -_{\mathbf{x} \in \mathcal{X}} \Bigg[\log p_{\mathcal{Z}}(\mathbf{z}) + \sum_{k=1}^{K}\log \Big| \operatorname{det} \Big(\frac{\partial \mathbf{h}_k}{\partial \mathbf{h}_{k - 1}}\Big) \Big| \Bigg]
\end{split}
\end{equation}
where $\mathbf{h}_k = f_1 \circ f_2 \circ .. \circ f_k(\mathbf{h}_0)$ with $\mathbf{h}_0 = \mathbf{x}$.

\subsection{Standard $n \times n$ Convolution} \label{sec:standard_nxn_conv}

In this section, we revisit the standard $n \times n$ convolution. Let $\mathbf{X}$ be an $C \times H \times W$ input; $\mathbf{W}$ is a $D \times C \times K$ kernel, and the convolution can be expressed as follows:
\begin{equation} \begin{split} \label{eqn:standard_nxn_conv}
    \mathbf{Y} &= \mathbf{W} \star \mathbf{X} 
               = \Big[{\mathbf{W}}_{:,:,1} \; {\mathbf{W}}_{:,:,2} \; \cdots \; {\mathbf{W}}_{:,:,k} \Big]
               \times 
               \begin{bmatrix} \mathbf{X}^1_{:,:,:} \\[5pt] \mathbf{X}^2_{:,:,:} \\[5pt] \vdots \\[5pt] \mathbf{X}^K_{:,:,:}\end{bmatrix} \\
               &= \sum_{k=1}^K {\mathbf{W}}_{:,:,k} \times \mathbf{X}^k_{:,:,:} = \sum_{k=1}^K {\mathbf{W}}_{:,:,k} \times \mathcal{S}_k(\mathbf{X}),
\end{split} \end{equation}
where $\mathbf{X}^k_{:,:,:}$ is a $C \times H \times W$ matrix that represents a spatially shifted version of input matrix $\mathbf{X}$ with shift amount $(i_k, j_k)$, . ${\mathbf{W}}_{:,:,k}$ represents the $D \times C$ matrix corresponding to the kernel index $k$, the symbol $\star$ denotes a convolution operator.

In Equation \eqref{eqn:standard_nxn_conv}, the standard convolution is simply a sum of $1 \times 1$ convolutions on shifted inputs. The function $\mathcal{S}_k$ maps the input $\mathbf{X}$ to the corresponding shifted input $\mathbf{X}^k_{:,:,:}$. The standard convolution uses the common shifted input with integer-valued shift amounts for index $k$. Figure \ref{fig:reformulated_conv} illustrates our reformulated $n \times n$ convolution, if we can share the shifted inputs regardless of the kernel index, especially $\mathcal{S}_k(\mathbf{X}) =\mathcal{S}(\mathbf{X})$, the standard convolution will be simplified as the $1 \times 1$ convolution as shown in Equation \ref{eqn:simplify_standard_nxn_conv}.
In this paper, we propose a shift function $\mathcal{S}$, which is an invertible and tractable form of the Jacobian determinant.
\begin{equation} \begin{split} \label{eqn:simplify_standard_nxn_conv}
    \sum_{k=1}^K {\mathbf{W}}_{:,:,k} \times \mathcal{S}_k(\mathbf{X}) &= \sum_{k=1}^K {\mathbf{W}}_{:,:,k} \times \mathcal{S}(\mathbf{X}) \\
    &= \Bigg(\sum_{k=1}^K {\mathbf{W}}_{:,:,k}\Bigg) \times \mathcal{S}(\mathbf{X}).
\end{split} \end{equation}

\begin{figure}[H]
    \includegraphics[width=0.95\columnwidth]{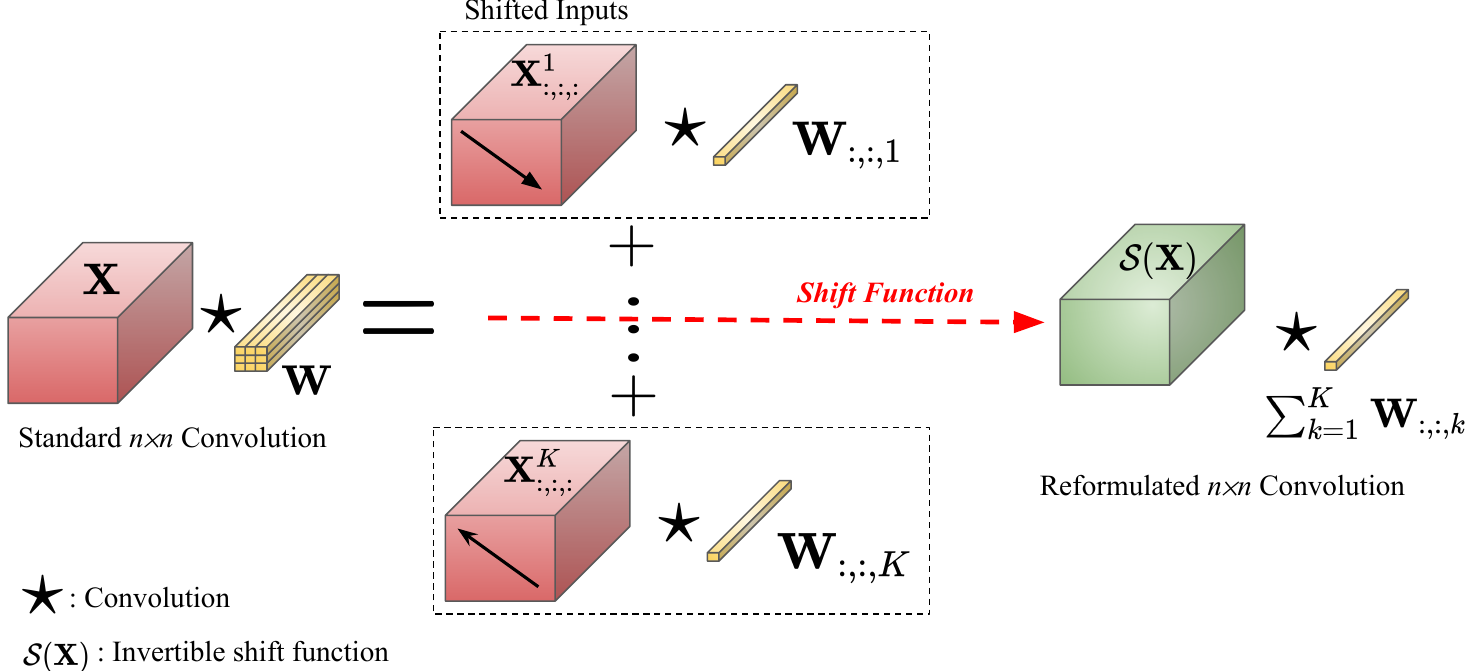}
    \caption{Reformulating $n \times n$ convolution. We propose to shift inputs instead of kernels. The proposed invertible $n \times n$ convolution will be simplified as a combination of the invertible shift function $\mathcal{S}$ and the invertible $1 \times 1$ convolution.}
    \label{fig:reformulated_conv}
\end{figure}

\section{Invertible $\boldmath{n} \times \boldmath{n}$ Convolution} \label{sec:the_proposed_method}
In this section, we first introduce our proposed Invertible Shift Function and then present invertible $n \times n$ convolution in details. 



\subsection{Invertible Shift Function}

The shift function $\mathcal{S}$ will approximate all shifted input $\mathbf{X}^k_{:,:,:}$ ($1 \leq k \leq K$). Here, we propose to design $\mathcal{S}$ as a linear transformation per channel; specifically, we have learnable variables $\alpha_c, \beta_c$; $1 \leq c \leq C$ are scale and translation parameters for each channel, respectively. The shift function $\mathcal{S}$ can be formulated as follows:
\begin{equation} \label{eqn:shift_function}
\mathcal{S}(\mathbf{X}_{c, i, j}) = \alpha_c\mathbf{X}_{c, i, j} + \beta_c,
\end{equation}
where $c, i, j$ are the depth channel index and spatial indices, respectively. The reverse function of $\mathcal{S}$ can be easy to obtain:
\begin{equation} \label{eqn:reverse_shift_function}
\mathbf{X}_{c, i, j} = \frac{\mathcal{S}(\mathbf{X}_{c, i, j}) - \beta_c}{\alpha_c}.
\end{equation}

Thanks to Equation \eqref{eqn:shift_function}, the value of $\mathcal{S}(\mathbf{X}_{c, i, j})$ only depends on $\mathbf{X}_{c, i, j}$ and the Jacobian matrix will be in the form of the diagonal matrix as follows:
\begin{equation} \label{eqn:jacobian_matrix}
\begin{split}
    \mathbf{J} = \frac{\partial \mathcal{S}(\mathbf{X})}{\partial\mathbf{X}} &= 
    \begin{bmatrix}
        \frac{\partial\mathcal{S}(\mathbf{X}_{1, 1, 1})}{\partial\mathbf{X}_{1, 1, 1}} & 0 & \cdots  &0 \\
        0 & \frac{\partial\mathcal{S}(\mathbf{X}_{1, 1, 2})}{\partial\mathbf{X}_{1, 1, 2}} & \cdots  &0 \\
        \vdots & \vdots  & \ddots & \vdots \\
        0 & 0 & \cdots & \frac{\partial\mathcal{S}(\mathbf{X}_{C, H, W})}{\partial\mathbf{X}_{C, H, W}} \\
    \end{bmatrix} \\
    &= 
    \begin{bmatrix}
        \alpha_1 & 0 & \cdots  &0 \\
        0 & \alpha_1& \cdots  &0 \\
        \vdots & \vdots  & \ddots & \vdots \\
        0 & 0 & \cdots & \alpha_c. \\
    \end{bmatrix}
\end{split}
\end{equation}

Therefore, the determinant of Equation \eqref{eqn:jacobian_matrix} is the product of all elements in the diagonal of the matrix $\mathbf{J}$ as in Equation \eqref{eqn:log_det}.
\begin{equation} \begin{split} \label{eqn:log_det}
        \operatorname{det}\Bigg(\frac{\partial \mathcal{S}(\mathbf{X})}{\partial\mathbf{X}}\Bigg) &= \prod_{c=1}^C \alpha_c^{H \times W} \\
        \log \Bigg|\operatorname{det}\Bigg(\frac{\partial \mathcal{S}(\mathbf{X})}{\partial\mathbf{X}}\Bigg)\Bigg| &= H \times W \times \sum_{c=1}^C \log |\alpha_c|.
\end{split} \end{equation}

\subsection{Invertible $n \times n$ Convolution}

Kingma \cite{glow} proposed invertible $1 \times 1$ convolution as the smart way to learn the permutation matrix instead of the fixed permutation \cite{nice, real_nvp}. However, the $1 \times 1$ suffers from limited flexibility compared to the standard convolution. In particular, the receptive fields of $1 \times 1$ convolution is limited. When the network goes deeper, the receptive fields of $1 \times 1$ convolutions are still small areas; these, therefore, cannot generalize or model large objects of high-dimensional data. 
However, the $1 \times 1$ convolution has its own advantages compared to the standard convolution. First, the $1 \times 1$ convolution allows the network to compress the data of the input volume to be smaller. Second, $1 \times 1$ suffers less over-fitting due to small kernel sizes. Therefore, in our proposal, we still take advantages of the $1 \times 1$ convolution. Specifically, we adopt the successfully invertible $1 \times 1$ convolution of Glow~\cite{glow} in our design.

In the previous subsection, we proved that the shift function $\mathcal{S}$ is invertible and proved the tractability of the Jacobian determinant. In Section \ref{sec:standard_nxn_conv}, we indicated that if we can share shifted inputs regardless of the kernel index via the shift function $\mathcal{S}$, we can simplify the standard $n \times n$ convolution to the composition of the $\mathcal{S}$ and $1 \times 1$ convolution. Therefore, the invertible $n \times n$ convolution will be equivalent to the combination of the invertible shift function $\mathcal{S}$ and the invertible $1 \times 1$ convolution. Specifically, the input will first be forwarded to the shift function $\mathcal{S}$ and then convoluted with the $1 \times 1$ filter. Algorithm \ref{alg:inv_nxn_conv} illustrates the pseudo code of the invertible $n \times n$ convolution.

\begin{algorithm}
\footnotesize
\SetAlgoLined
\textbf{Input:} An input $\mathbf{X} \in \mathbb{R}^{N \times H \times W \times C}$

\KwResult{An output of invertible $n \times n$ convolution and the log Jacobian determinant}

Initialize $\alpha, \beta \in \mathbb{R}^{C}$ for the invertible shift function\;
Initialize $\mathbf{W} \in \mathbb{R}^{C \times C}$ as a rotation matrix for the invertible $1 \times 1$ convolution function\;
logdet = 0.0\;

\textbf{\textit{The invertible shift function}}\;
$\mathbf{Y} = \mathbf{X} \times \alpha + \beta$\ (Channel-wise operations)\; 
The inverse will be $\mathbf{X} = \frac{\mathbf{Y} - \beta}{\alpha}$\;
logdet = logdet + $\sum_{i=1}^C\log(\alpha_i)$\;

\textbf{\textit{The invertible $1 \times 1$ convolution}}\;
$\mathbf{Z} = Conv(\mathbf{Y}, \mathbf{W})$\;
The inverse will be $\mathbf{Y} = Conv(\mathbf{Z}, \mathbf{W}^{-1}$)\;
logdet = logdet + $\log(\operatorname{det}(\mathbf{W}))$ * $H$ * $W$\;

Return $\mathbf{Z}$ and logdet\;
\caption{Invertible $n \times n$ Convolution}
\label{alg:inv_nxn_conv}
\end{algorithm}

Figure \ref{fig:invertible_nxn_conv}a illustrates our one step of flow. We adopt the common design of a flow step~\cite{glow, dxd_invertible_conv, Truong_EUNVP} in our design. Our proposal can be easily integrated to the multi-scale architecture designed by Dinh et~al. \cite{real_nvp} (Figure \ref{fig:invertible_nxn_conv}b). From our proposal, we can generalize the invertible $1 \times 1$ convolution to the invertible $n \times n$ convolution through the shift function $\mathcal{S}$. It can help to encourage the filters to learn a more efficient data representation and embed more useful latent features than the invertible $1 \times 1$ convolution used in Glow \cite{glow}. Besides, we use fewer parameters and have less inference time compared to the standard $n \times n$ convolutions.

\vspace{-6pt} 
\begin{figure}[H]
    \includegraphics[width=0.7\columnwidth]{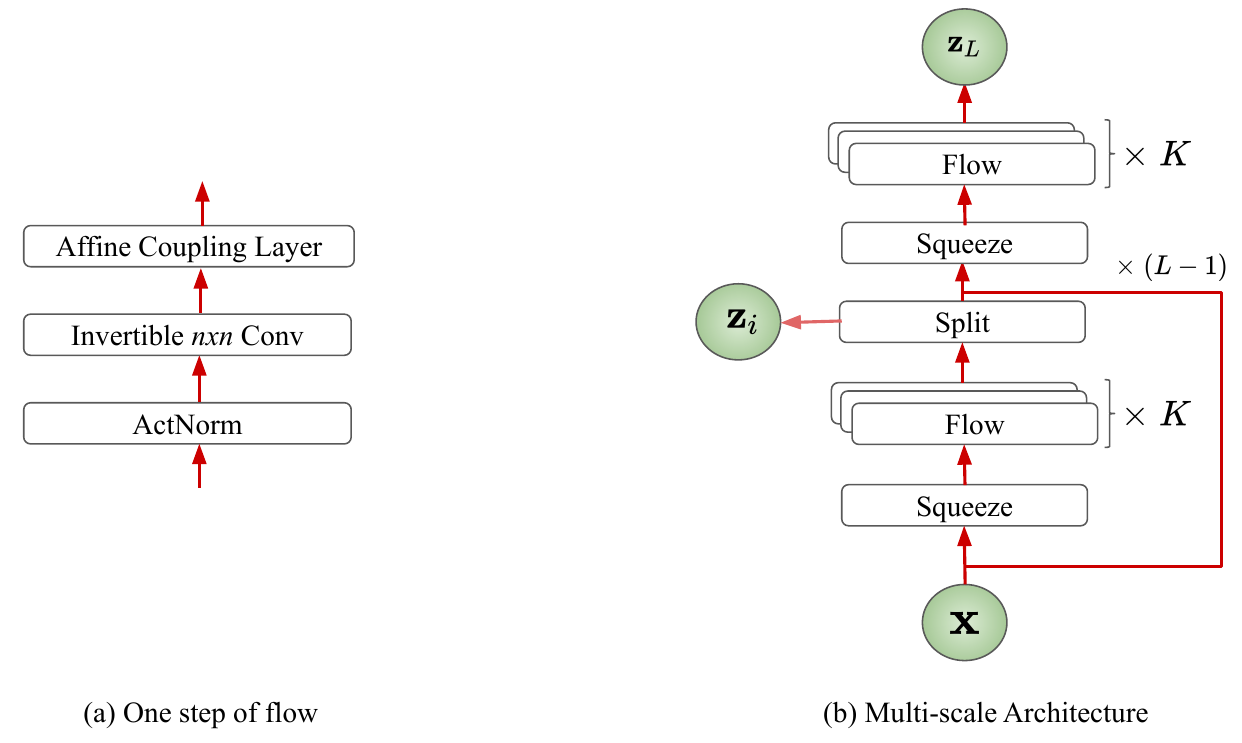}
    \caption{(\textbf{a}) is our one step of flow using an invertible $n \times n$ convolution. Our proposal flow step is able to combine with the multi-scale architecture designed in RealNVP (\textbf{b}). $K$ and $L$ are the depth of flow and the number of levels, respectively.}
    \label{fig:invertible_nxn_conv}
\end{figure}

\section{Experiments} \label{sec:experiments}

In this section, we present our experimental results on CIFAR-10, ImageNet and Celeb-HQ datasets. Firstly, in Section \ref{sec:quantitative_exp}, we compare log-likelihood against the previous flow-based models, that is, RealNVP \cite{real_nvp}, Glow \cite{glow} and Emerging Convolution \cite{dxd_invertible_conv}. Finally, in Section \ref{sec:quanlitative_exp}, we show our qualitative results trained on the Celeb-HQ dataset.

\subsection{Quantitative Experiments} \label{sec:quantitative_exp}

\noindent
\textbf{Datasets and Metric: }
We evaluate our invertible $n \times n$ convolution on CIFAR-10 \mbox{(Figure~\ref{fig:imagenet_cifar}a)} and ImageNet \mbox{(Figure~\ref{fig:imagenet_cifar}b)} with $32 \times 32$ and $64 \times 64$ image sizes. We use bits per dimension as the criteria with which to evaluate models. We compare our method against RealNVP \cite{real_nvp}, Glow \cite{glow} and Emerging Convolution \cite{dxd_invertible_conv}. We adopt the network structures of Glow and replace all invertible $1 \times 1$ convolutions of Glow with our invertible $n \times n$ convolutions. For the data preprocessing, we follow the same process as in RealNVP \cite{real_nvp}.

\end{paracol}
\nointerlineskip
\begin{figure}[H]
    \widefigure
    \includegraphics[width=0.9\columnwidth]{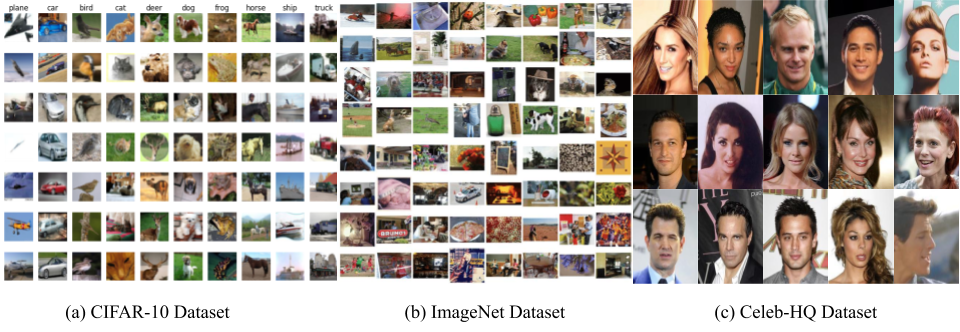}
    \caption{The examples from the CIFAR dataset (\textbf{a}), ImageNet dataset (\textbf{b}) and Celeb-HQ dataset (\textbf{c})}
    \label{fig:imagenet_cifar}
\end{figure}
\begin{paracol}{2}
\switchcolumn

\noindent
\textbf{Network Configurations:} In the CIFAR experiment, the depth of flow $K$ and the number of levels $L$ are set to $32$ and $3$, respectively. Meanwhile, the depth of flow in ImageNet experiments is set to $48$, the numbers of levels of ImageNet $32 \times 32$ and ImageNet $64 \times 64$ experiments are set to $3$ and $4$, respectively. We use the Adam optimizer \cite{adam_optim} to optimize the networks in which batch size and learning rate are set to $64$ (per GPU) and $0.001$, respectively. We choose Normal Distribution as the prior distribution $p_{\mathcal{Z}}(z) \sim \mathcal{N}(\mathbf{z}; 0, \mathbf{I})$ in all~experiments.

\textls[-15]{The shift function $\mathcal{S}$ will be not inverse if the $\alpha_c = 0$ ($\exists\;c \in [1...C]$). Hence, in the training process, we will first initialize $\alpha_c = 1 \text{ and } \beta_c = 0$ ($1 \leq c \leq C$). During the learning processing, we keep $\alpha_c$ ($1 \leq c \leq C$) as a different $0$ to guarantee that the shift function $\mathcal{S}$ is inverse and to guarantee the tractability of the Jacobian determinant. 
Training models on high-dimensional data requires large memory. To be able to train with a large batch size, we simultaneously and distributively trained the models on four GPUs via
Horovod \mbox{(\url{https://github.com/horovod/horovod})} and TensorFlow (\url{https://tensorflow.org}) frameworks. } 


\noindent
\textbf{Results:} Table \ref{tab:bit_per_dims} shows our experimental results. 
In particular, our proposal helps to improve the generative models on ImageNet $32 \times 32$ and ImageNet $64 \times 64$ datasets, which are more challenging than CIFAR-10.
In particular, our proposed method achieves a state-of-the-art performance in which the bit per dimension results in ImageNet \mbox{$32 \times 32$}, and ImageNet \mbox{$64 \times 64$} is 
\textbf{3.96} and \textbf{3.74}, respectively.
In comparison, the Emerging Convolution~\cite{dxd_invertible_conv} and Glow achieve similar results in both ImageNet \mbox{$32 \times 32$} and ImageNet \mbox{$64 \times 64$} benchmarks, which are 4.09 and 3.81, respectively. Meanwhile, the corresponding results of RealNVP on these benchmarks are 4.28 and 3.98, respectively. 
As shown by the results, our proposed invertible $n \times n$ convolution provides a better generative capability than the stand-alone invertible $1 \times 1$ convolution. 
Since Emerging Convolution uses invertible auto-regressive convolution, our proposal is, therefore, less complicated and has faster inference than Emerging Convolution. 
In the CIFAR-10 benchmark, although our model does not perform as well as  Glow \cite{glow} and Emerging Convolution \cite{dxd_invertible_conv}, we find it interesting that our method gains competitive results with a small number of modifications. The gap in performance is partially caused by the small amount of CIFAR-10 data that is inefficient for training the well-generalized convolution.

\begin{specialtable}[H]
    \small
    \caption{Comparative results (bits per dimension) of proposed invertible $n \times n$ convolution compared to RealNVP, Glow and Emerging Convolution.}
    \begin{tabular}{cccc}
         \toprule
         \textbf{\,\,\,\,\,\,\,\,\,\,\,\,\,Models\,\,\,\,\,\,\,\,\,\,\,\,\,}   & \textbf{\,\,\,\,\,\,\,\,\,\,\,\,\,\,\,CIFAR-10\,\,\,\,\,\,\,\,\,\,\,\,\,\,\,} & \textbf{\,\,\,\,\,\,\,\,\,\,\,\,\,\,ImageNet 32\,\,\,\,\,\,\,\,\,\,\,\,\,\,} & \textbf{\,\,\,\,\,\,\,\,\,\,\,\,\,\,\,ImageNet 64\,\,\,\,\,\,\,\,\,\,\,\,\,\,\,} \\
         \midrule
         RealNVP        & 3.49 & 4.28 & 3.98 \\
         Glow           & 3.35 & 4.09 & 3.81 \\
         Emerging Conv  & \textbf{3.34} & 4.09 & 3.81 \\
         \textbf{Ours}  & 3.50 & \textbf{3.96} & \textbf{3.74}\\
         \bottomrule
    \end{tabular}
    \label{tab:bit_per_dims}
\end{specialtable}

\subsection{Qualitative Experiments} \label{sec:quanlitative_exp}

The CelebA-HQ dataset \cite{celeb_dataset} was selected to train the model using the architectures defined in the previous section with a higher resolution ($256 \times 256$ image sizes). The depth of flow $K$ and the number of levels $L$ were set to $32$ and $6$, respectively. Since high-dimensional data requires large memory, we reduced the batch size to $1$ (per GPU) and trained on eight GPUs. 
The qualitative experiment aims to study the efficiency of the model when it scales up to the high-resolution images, synthesizes realistic images, and provides the meaningful latent space.
Figure \ref{fig:imagenet_cifar}c shows the examples of Celeb-HQ datasets. We trained our model on 5-bit images in order to improve visual quality with a slight trade-off of color fidelity.
As shown by the synthetic images in Figure \ref{fig:syn_celeb_img}, our model can generalize realistic images in high dimensional data.

\end{paracol}
\nointerlineskip
\begin{figure}[H]
    \widefigure
    \includegraphics[width=0.95\columnwidth]{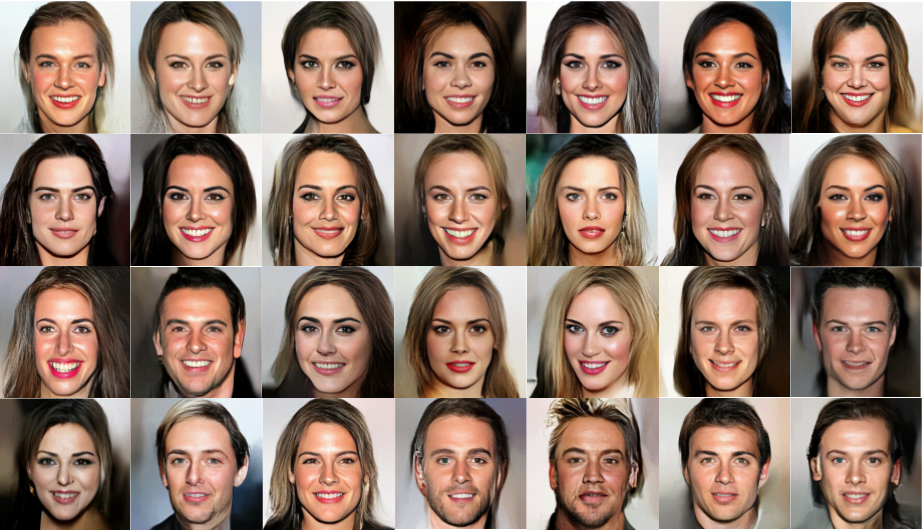}
    \caption{Synthetic celebrity faces sampled from our model trained on the CelabA-HQ dataset.}
    \label{fig:syn_celeb_img}
\end{figure}
\begin{paracol}{2}
\switchcolumn

\vspace{-10pt} 

\section{Conclusions and Future Work} \label{sec:label}
This paper has presented a novel invertible $n \times n$ convolution approach. By reformulating the convolution layer, we propose to use the shift function to shift inputs instead of kernels. We prove that our shift function is invertible and tractable in terms of calculating the Jacobian determinant. The method leverages the shift function and the invertible $1 \times 1$ convolution to generalize to the invertible $n \times n$ convolution. Through experiments, our proposal has achieved state-of-the-art results in quantitative measurement and is able to generate realistic images with high-resolution.

There are several challenges that remain to be addressed in future work. In particular, when the model scales up to the high-resolution images, it requires a large amount of GPU memory during the training process, that is, the back-propagation process. Maintaining the rotation matrix property for the invertible $1 \times 1$ convolution when training the model on a large dataset is also a challenging task, since the model easily falls into the non-inverse matrix due to the stochastic gradient update of the back-propagation algorithm. That issue is interesting work and should be improved in the future.


\vspace{6pt} 



\authorcontributions{
Conceptualization: T.-D.T. and C.N.D. Methodology: T.-D.T., C.N.D. and K.L. Review and Editing: K.L., M.-T.T., and N.L. Supervision: K.L. and M.-T.T. All authors have read and agreed to the published version of the manuscript.
}

\dataavailability{CIFAR Dataset \url{https://www.cs.toronto.edu/~kriz/cifar.html}, ImageNet dataset \url{https://image-net.org/}, and CelebA-HQ Dataset \url{https://mmlab.ie.cuhk.edu.hk/projects/CelebA.html}.} 


\conflictsofinterest{The authors declare no conflict of interest.
}

\end{paracol}
\reftitle{References}


\begin{thebibliography}{999}

\bibitem[He \em{et~al.}(2016)He, Zhang, Ren, and Sun]{he2016deep}
He, K.; Zhang, X.; Ren, S.; Sun, J. Deep residual learning for image recognition. In Proceedings of the IEEE Conference on Computer Vision and Pattern
  Recognition,  {Las Vegas, NV, USA, 27--30 June 2016;} pp. 770--778.

\bibitem[Huang \em{et~al.}(2017)Huang, Liu, Van Der~Maaten, and
  Weinberger]{huang2017densely}
Huang, G.; Liu, Z.; Van Der~Maaten, L.; Weinberger, K.Q.
\newblock Densely connected convolutional networks.
\newblock  In Proceedings of the IEEE Conference on Computer Vision and Pattern
  Recognition,  {Honolulu, HI, USA, 21--26 July 2017;} pp. 4700--4708.

\bibitem[Sun \em{et~al.}(2018)Sun, Pang, Shi, Yi, and Ouyang]{sun2018fishnet}
Sun, S.; Pang, J.; Shi, J.; Yi, S.; Ouyang, W.
\newblock FishNet: A Versatile Backbone for Image, Region, and Pixel Level
  Prediction.
\newblock  \emph{Adv. Neural Inf. Proc. Syst.}  {{2018}}, 760--770.

\bibitem[Liu \em{et~al.}(2016)Liu, Anguelov, Erhan, Szegedy, Reed, Fu, and
  Berg]{liu2016ssd}
Liu, W.; Anguelov, D.; Erhan, D.; Szegedy, C.; Reed, S.; Fu, C.Y.; Berg, A.C.
\newblock Ssd: Single shot multibox detector.
\newblock  In European Conference on Computer Vision, {Amsterdam, The Netherlands, 8--16 October 2016}; Springer:  {Berlin/Heidelberg, Germany,} 
  2016;~pp. 21--37.

\bibitem[Redmon \em{et~al.}(2016)Redmon, Divvala, Girshick, and
  Farhadi]{redmon2016you}
Redmon, J.; Divvala, S.; Girshick, R.; Farhadi, A.
\newblock You only look once: Unified, real-time object detection.
\newblock  In Proceedings of the IEEE Conference on Computer Vision and Pattern
  Recognition,  {Las Vegas, NV, USA, 27--30 June 2016;} pp. 779--788.

\bibitem[Lin \em{et~al.}(2017)Lin, Goyal, Girshick, He, and
  Doll{\'a}r]{lin2017focal}
Lin, T.Y.; Goyal, P.; Girshick, R.; He, K.; Doll{\'a}r, P.
\newblock Focal loss for dense object detection.
\newblock  In Proceedings of the IEEE International Conference on Computer Vision, V{enice, Italy
  22--29 October 2017;} pp. 2980--2988.
  
\bibitem{Luu_IJCB2011}Luu, K., Seshadri, K., Savvides, M., Bui, T. \& Suen, C. Contourlet Appearance Model for Facial Age Estimation. {\em Intl. Joint Conf. On Biometrics (IJCB)}. pp. 1-7 (2011)

\bibitem{Le_JPR2015}Le, H., Seshadri, K., Luu, K. \& Savvides, M. Facial Aging and Asymmetry Decomposition Based Approaches to Identiﬁcation of Twins. {\em Journal Of Pattern Recognition}. \textbf{48} pp. 3843-3856 (2015)

\bibitem{Xu_TIP2015}Xu, F., Luu, K. \& Savvides, M. Spartans: Single-sample Periocular-based Alignment-robust Recognition Technique Applied to Non-frontal Scenarios. {\em Trans. On Image Processing (TIP)}. \textbf{24} pp. 4780-4795 (2015)

\bibitem{Xu_IJCB2011}Xu, J., Luu, K., Savvides, M., Bui, T. \& Suen, C. Investigating Age Invariant Face Recognition Based on Periocular Biometrics. {\em Intl. Joint Conf. On Biometrics (IJCB)}. (2011)

\bibitem{Duong_ICASSP2011}Duong, C., Quach, K., Luu, K., Le, H. \& Jr, K. Fine Tuning Age Estimation with Global and Local Facial Features. {\em Intl. Conf. On Acoustics, Speech And Signal Processing (ICASSP)}. pp. 1-7 (2011)

\bibitem{Luu_BTAS2009}Luu, K., Bui, T., Jr., K. \& Suen, C. Age Estimation using Active Appearance Models and Support Vector Machine Regression. {\em Intl. Conference On Biometrics: Theory, Applications And Systems (BTAS)}. (2009)

\bibitem{Luu_FG2011}Luu, K., Bui, T. \& Suen, C. Kernel Spectral Regression of Perceived Age from Hybrid Facial Features. {\em Conf. On Automatic Face And Gesture Recognition (FG)}. pp. 1-7 (2011)

\bibitem{Chen_FG2011}Chen, C., Yang, W., Wang, Y., Ricanek, K. \& Luu, K. Facial Feature Fusion and Model Selection for Age Estimation. {\em Conf. On Automatic Face And Gesture Recognition (FG)}. pp. 1-7 (2011)

\bibitem{Luu_ROBUST2008}Luu, K., Jr., K., Bui, T. \& Suen, C. The Familial Face Database: A Longitudinal Study of Family-based Growth and Development on Face Recognition. {\em Robust Biometrics: Understanding Science \& Technology (ROBUST)}. (2008)

\bibitem{Luu_CAI2011}Luu, K. Computer Approaches for Face Aging Problems. {\em The 23th Canadian Conference On Artificial Intelligence (CAI)}. (2010)

\bibitem[Chen \em{et~al.}(2017)Chen, Papandreou, Kokkinos, Murphy, and
  Yuille]{chen2017deeplab}
Chen, L.C.; Papandreou, G.; Kokkinos, I.; Murphy, K.; Yuille, A.L.
\newblock Deeplab: Semantic image segmentation with deep convolutional nets,
  atrous convolution, and fully connected crfs.
\newblock {\em IEEE Trans. Pattern Anal. Mach. Intell.}
  {\bf 2017}, {\em 40},~834--848.

\bibitem[Long \em{et~al.}(2015)Long, Shelhamer, and Darrell]{long2015fully}
Long, J.; Shelhamer, E.; Darrell, T.
\newblock Fully convolutional networks for semantic segmentation.
\newblock  In Proceedings of the IEEE Conference on Computer Vision and Pattern
  Recognition, {Boston, MA, USA, 7--12 June 2015;} pp. 3431--3440.

\bibitem[Mirza and Osindero(2014)]{mirza2014conditional}
Mirza, M.; Osindero, S.
\newblock Conditional generative adversarial nets.
\newblock {\em arXiv} {\bf 2014}, arXiv:1411.1784

\bibitem[Karras \em{et~al.}(2017)Karras, Aila, Laine, and
  Lehtinen]{karras2017progressive}
Karras, T.; Aila, T.; Laine, S.; Lehtinen, J.
\newblock Progressive growing of gans for improved quality, stability, and
  variation.
\newblock {\em arXiv} {\bf 2017}, arXiv:1710.10196

\bibitem{Duong_Long}Duong, C., Luu, K., Quach, K. \& Bui, T. Longitudinal Face Modeling via Temporal Deep Restricted Boltzmann Machines. {\em 2016 IEEE Conference On Computer Vision And Pattern Recognition (CVPR)}. pp. 5772-5780 (2016)

\bibitem[Mattia \em{et~al.}(2019)Mattia, Galeone, Simoni, and
  Ghelfi]{dimattia2019survey}
Mattia, F.D.; Galeone, P.; Simoni, M.D.; Ghelfi, E.
\newblock A Survey on GANs for Anomaly Detection. {\em arXiv} \textbf{2019}, arXiv:cs.LG/1906.11632.
 

\bibitem[Yu \em{et~al.}(2018)Yu, Lin, Yang, Shen, Lu, and
  Huang]{yu2018generative}
Yu, J.; Lin, Z.; Yang, J.; Shen, X.; Lu, X.; Huang, T.S.
\newblock Generative image inpainting with contextual attention.
\newblock  In Proceedings of the IEEE Conference on Computer Vision and Pattern
  Recognition, {Salt Lake City, UT, USA, 18--23 June 2018;} pp. 5505--5514.

\bibitem[Karras \em{et~al.}(2019)Karras, Laine, and Aila]{karras2019style}
Karras, T.; Laine, S.; Aila, T.
\newblock A style-based generator architecture for generative adversarial
  networks.
\newblock  In Proceedings of the IEEE/CVF Conference on Computer Vision and
  Pattern Recognition, {Long Beach, CA, USA, 15--20 June 2019}; pp. 4401--4410.

\bibitem[Ledig \em{et~al.}(2017)Ledig, Theis, Husz{\'a}r, Caballero,
  Cunningham, Acosta, Aitken, Tejani, Totz, Wang, et~al.]{ledig2017photo}
Ledig, C.; Theis, L.; Husz{\'a}r, F.; Caballero, J.; Cunningham, A.; Acosta,
  A.; Aitken, A.; Tejani, A.; Totz, J.; Wang, Z.; et al.
\newblock Photo-realistic single image super-resolution using a generative
  adversarial network.
\newblock  In Proceedings of the IEEE Conference on Computer Vision and Pattern
  Recognition, {Honolulu, HI, USA, 21--26 July 2017;} pp. 4681--4690.
  
  
\bibitem{Duong_TNVP}Duong, C., Quach, K., Luu, K., Le, T. \& Savvides, M. Temporal Non-volume Preserving Approach to Facial Age-Progression and Age-Invariant Face Recognition. {\em 2017 IEEE International Conference On Computer Vision (ICCV)}. pp. 3755-3763 (2017)

\bibitem{Duong_automatic}Duong, C., Luu, K., Quach, K., Nguyen, N., Patterson, E., Bui, T. \& Le, N. Automatic Face Aging in Videos via Deep Reinforcement Learning. {\em 2019 IEEE/CVF Conference On Computer Vision And Pattern Recognition (CVPR)}. pp. 10005-10014 (2019)

\bibitem{Duong_2018}Duong, C., Luu, K., Quach, K. \& Bui, T. Deep Appearance Models: A Deep Boltzmann Machine Approach for Face Modeling. {\em International Journal Of Computer Vision}. \textbf{127}, 437-455 (2018,8), http://dx.doi.org/10.1007/s11263-018-1113-3

\bibitem[Kingma and Dhariwal(2018)]{glow}
Kingma, D.P.; Dhariwal, P.
\newblock Glow: Generative Flow with Invertible 1x1 Convolutions. In {\em
  Advances in Neural Information Processing Systems 31}; Bengio, S., Wallach,
  H., Larochelle, H., Grauman, K., Cesa-Bianchi, N., Garnett, R., Eds.; Curran
  Associates, Inc.: {Red Hook, NY, USA,} 2018; pp. 10215--10224.

\bibitem[Kingma \em{et~al.}(2016)Kingma, Salimans, Jozefowicz, Chen, Sutskever,
  and Welling]{invertible_autoregressive}
Kingma, D.P.; Salimans, T.; Jozefowicz, R.; Chen, X.; Sutskever, I.; Welling,
  M.
\newblock Improved Variational Inference with Inverse Autoregressive Flow. In
  {\em Advances in Neural Information Processing Systems 29}; Lee, D.D.,
  Sugiyama, M., Luxburg, U.V., Guyon, I., Garnett, R., Eds.; Curran Associates,
  Inc.: {Red Hook, NY, USA,} 2016; pp. 4743--4751.

\bibitem{Truong_EUNVP}Truong, D., Nhan Duong, C., Luu, K., Tran, M. \& Le, N. Domain Generalization via Universal Non-volume Preserving Approach. {\em 2020 17th Conference On Computer And Robot Vision (CRV)}. pp. 93-100 (2020)


\bibitem[Dinh \em{et~al.}(2015)Dinh, Krueger, and Bengio]{nice}
Dinh, L.; Krueger, D.; Bengio, Y.
\newblock NICE: Non-linear Independent Components Estimation. \emph{{arXiv}} 2015, {arXiv:1410.8516}. 

\bibitem[Dinh \em{et~al.}(2017)Dinh, Sohl-Dickstein, and Bengio]{real_nvp}
Dinh, L.; Sohl-Dickstein, J.; Bengio, S.
\newblock Density estimation using Real NVP.
\newblock   In Proceedings of the 3rd International Conference on Learning Representations, {ICLR}, {Toulon, France, 24--26 April 2017.}

\bibitem[Goodfellow \em{et~al.}(2014)Goodfellow, Pouget-Abadie, Mirza, Xu,
  Warde-Farley, Ozair, Courville, and Bengio]{gan}
Goodfellow, I.; Pouget-Abadie, J.; Mirza, M.; Xu, B.; Warde-Farley, D.; Ozair,
  S.; Courville, A.; Bengio, Y.
\newblock Generative Adversarial Nets. In {\em Advances in Neural Information
  Processing Systems 27}; Ghahramani, Z., Welling, M., Cortes, C., Lawrence,
  N.D., Weinberger, K.Q., Eds.; Curran Associates, Inc.: {Red Hook, NY, USA,} 2014; pp. 2672--2680.v

\bibitem[Kingma and Welling(2014)]{variational_autoencoder}
Kingma, D.P.; Welling, M.
\newblock Auto-Encoding Variational Bayes.
\newblock  In Proceedings of the 2nd International Conference on Learning Representations, {ICLR}
  2014, Banff, AB, Canada, 14--16 April 2014. {Conference Track Proceedings,
  2014.}

\bibitem[Hoogeboom \em{et~al.}(2019)Hoogeboom, van~den Berg, and
  Welling]{dxd_invertible_conv}
Hoogeboom, E.; van~den Berg, R.; Welling, M.
\newblock Emerging Convolutions for Generative Normalizing Flows. {\em {CoRR}} {\bf 2019}, {\em abs/1901.11137}. 

\bibitem[Papamakarios \em{et~al.}(2017)Papamakarios, Murray, and
  Pavlakou]{mask_autoregessive}
Papamakarios, G.; Murray, I.; Pavlakou, T.
\newblock Masked Autoregressive Flow for Density Estimation. In {\em Advances
  in Neural Information Processing Systems 30}; Guyon, I., Luxburg, U.V.,
  Bengio, S., Wallach, H., Fergus, R., Vishwanathan, S., Garnett, R., Eds.,
  Curran Associates, Inc.: {Red Hook, NY, USA,} 2017; pp. 2335--2344.

\bibitem[Behrmann \em{et~al.}(2019)Behrmann, Grathwohl, Chen, Duvenaud, and
  Jacobsen]{i-resnet}
Behrmann, J.; Grathwohl, W.; Chen, R.T.Q.; Duvenaud, D.; Jacobsen, J.H.
\newblock Invertible Residual Networks.
\newblock  In Proceedings of the 36th International Conference on Machine
  Learning, {Beach, CA, USA, 10--15 June 2019}; Chaudhuri, K., Salakhutdinov, R., Eds.; PMLR: Long Beach,
  California, USA, 2019; Volume~97; pp. 573--582.

\bibitem[Kim \em{et~al.}(2021)Kim, Papamakarios, and Mnih]{inv_self_attention}
Kim, H.; Papamakarios, G.; Mnih, A.
\newblock The Lipschitz Constant of Self-Attention. \emph{arXiv} 2021, arXiv:stat.ML/2006.04710


\bibitem[Chen \em{et~al.}(2019)Chen, Behrmann, Duvenaud, and
  Jacobsen]{residual_flows}
Chen, R.T.; Behrmann, J.; Duvenaud, D.; Jacobsen, J.H.
\newblock Residual flows for invertible generative modeling.
\newblock {\em arXiv} {\bf 2019}, arXiv:1906.02735.

\bibitem[Krizhevsky(2009)]{cifar}
Krizhevsky, A.
\newblock {Learning Multiple Layers of Features from Tiny Images.} {\bf 2009}.

\bibitem[Deng \em{et~al.}(2009)Deng, Dong, Socher, Li, Li, and
  Fei-Fei]{imagenet_cvpr09}
Deng, J.; Dong, W.; Socher, R.; Li, L.J.; Li, K.; Fei-Fei, L.
\newblock {ImageNet: A Large-Scale Hierarchical Image Database}.
\newblock  In Proceedings of Computer Vision and Pattern Recognition (CVPR), {Miami, FL, USA, 20--25 June 2009.}


\bibitem[Liu \em{et~al.}(2015)Liu, Luo, Wang, and Tang]{celeb_dataset}
Liu, Z.; Luo, P.; Wang, X.; Tang, X.
\newblock Deep Learning Face Attributes in the Wild.
\newblock  In Proceedings of International Conference on Computer Vision (ICCV), {Santiago, Chile, 7--13 December 2015.}

\bibitem[Ho \em{et~al.}(2019)Ho, Chen, Srinivas, Duan, and Abbeel]{flow_pp}
Ho, J.; Chen, X.; Srinivas, A.; Duan, Y.; Abbeel, P.
\newblock Flow++: Improving Flow-Based Generative Models with Variational
  Dequantization and Architecture Design. \emph{arXiv} \textbf{2019}, arXiv:cs.LG/1902.00275


\bibitem[Vaswani \em{et~al.}(2017)Vaswani, Shazeer, Parmar, Uszkoreit, Jones,
  Gomez, Kaiser, and Polosukhin]{transformer}
Vaswani, A.; Shazeer, N.; Parmar, N.; Uszkoreit, J.; Jones, L.; Gomez, A.N.;
  Kaiser, L.U.; Polosukhin, I.
\newblock Attention is All you Need.
\newblock  In \emph{Advances in Neural Information Processing Systems}; Guyon, I.;
  Luxburg, U.V.; Bengio, S.; Wallach, H.; Fergus, R.; Vishwanathan, S.;
  Garnett, R., Eds.; Curran Associates, Inc.: {Red Hook, NY, USA,} 2017; Volume~30.

\bibitem[Germain \em{et~al.}(2015)Germain, Gregor, Murray, and
  Larochelle]{made}
Germain, M.; Gregor, K.; Murray, I.; Larochelle, H.
\newblock MADE: Masked Autoencoder for Distribution Estimation.
\newblock  In Proceedings of the 32nd International Conference on Machine
  Learning, Lille, France, 6--11 July 2015; Bach, F.; Blei, D., Eds.; PMLR: Lille, France, 2015; Volume~37, pp. 881--889.

\bibitem[Kingma and Ba(2015)]{adam_optim}
Kingma, D.P.; Ba, J.
\newblock Adam: {A} Method for Stochastic Optimization.
\newblock In Proceedings of the 3rd International Conference on Learning Representations, {ICLR}
  2015, San Diego, CA, USA, May 7--9, 2015. {Conference Track Proceedings, 2015.}






\end{thebibliography}
\end{document}